\documentclass[conference]{IEEEtran}
\IEEEoverridecommandlockouts
\usepackage{cite}
\usepackage{amsmath,amssymb,amsfonts}
\usepackage{algorithmic}
\usepackage{graphicx}
\usepackage{textcomp}
\usepackage{xcolor}
\usepackage{tikz}

\usepackage[caption=false]{subfig}  
\usepackage{amstext}
\usepackage{booktabs}
\usepackage{multirow}
\usepackage[backref=page]{hyperref}

\hypersetup{pdftitle={velioglu2024fashionfail}}
\usepackage{cleveref}

\newcommand{\eg}{e.g.\ }
\newcommand{\cf}{cf.\ }

\newcommand{\opacity}{0.9}  
\Crefname{figure}{Fig.}{Figs.}

\usepackage{pifont}
\newcommand{\cmark}{\ding{51}}
\newcommand{\xmark}{\ding{55}}

\begin{document}

\title{FashionFail: Addressing Failure Cases \\in Fashion Object Detection and Segmentation\thanks{This work has been funded by the German federal state of North Rhine-Westphalia as part of the research training group ``Dataninja'' (Trustworthy AI for Seamless Problem Solving: Next Generation Intelligence Joins Robust Data Analysis) and KI-Starter (grant no. 005-2204-0023).}}

\author{\IEEEauthorblockN{
    Riza Velioglu, 
    Robin Chan, and 
    Barbara Hammer 
    }
    \IEEEauthorblockA{\textit{Machine Learning Group, CITEC}
    \textit{Bielefeld University},
    Bielefeld, Germany \\
    \{rvelioglu, rchan, bhammer\}@techfak.de}}

© 2024 IEEE. Personal use of this material is permitted. Permission from IEEE must be obtained for all other uses, in any current or future media, including reprinting/republishing this material for advertising or promotional purposes, creating new collective works, for resale or redistribution to servers or lists, or reuse of any copyrighted component of this work in other works.

\maketitle

\begin{abstract}
In the realm of fashion object detection and segmentation for online shopping images, existing state-of-the-art fashion parsing models encounter limitations, particularly when exposed to non-model-worn apparel and close-up shots. To address these failures, we introduce \emph{FashionFail}; a new fashion dataset with e-commerce images for object detection and segmentation. The dataset is efficiently curated using our novel annotation tool that leverages recent foundation models.
The primary objective of FashionFail is to serve as a test bed for evaluating the robustness of models.
Our analysis reveals the shortcomings of leading models, such as Attribute-Mask R-CNN and Fashionformer.
Additionally, we propose a baseline approach using naive data augmentation to mitigate common failure cases and improve model robustness.
Through this work, we aim to inspire and support further research in fashion item detection and segmentation for industrial applications. 
The dataset, annotation tool, code, and models are available at \url{https://rizavelioglu.github.io/fashionfail/}.
\end{abstract}


\section{Introduction}
\label{sec:introduction}

Computer vision has impacted many fields across different industries, such as autonomous driving \cite{janai2020computer,geiger2012we}, health care \cite{gao2018computer,esteva2021deep} or retail \cite{wei2020deep,shankar2021technology}. As fashion is undoubtedly one of the largest industries in the world, this field has also become a popular research field enabling many applications, including recommendation systems \cite{Veit_2017_CVPR,Lin_2020_CVPR,Hou_2021_ICCV} or outfit synthesis \cite{han2018viton,Yang_2020_CVPR,zhu2017your}.

All methods used in fashion applications share a common reliance on successful fashion item detection. What poses a particular challenge in recognizing fashion items, as evidenced in Fashionpedia images~\cite{jia2020fashionpedia}, is the vast diversity in garment appearances and the variability in poses exhibited by individuals wearing them. Moreover, the presence of cluttered backgrounds in certain images further compounds the difficulty of the detection task. Nevertheless, recent progress has yielded remarkable results for state-of-the-art models in accurately segmenting (and therefore detecting) fashion items in complex scenes~\cite{jia2020fashionpedia,xu2022fashionformer}. These models have demonstrated the ability to reliably segment individual garment instances~(\eg jackets) along with their respective garment parts (\eg collar, sleeve, pocket, buttons, ...).

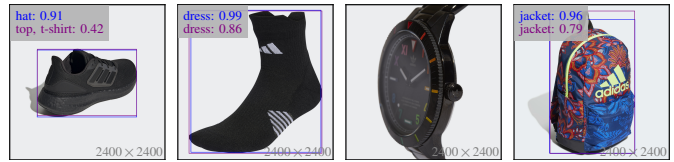
\begin{figure}
    \centering
    \begin{tikzpicture}
        \node[inner sep=0pt] at (0,0) {\includegraphics[width=.99\linewidth]{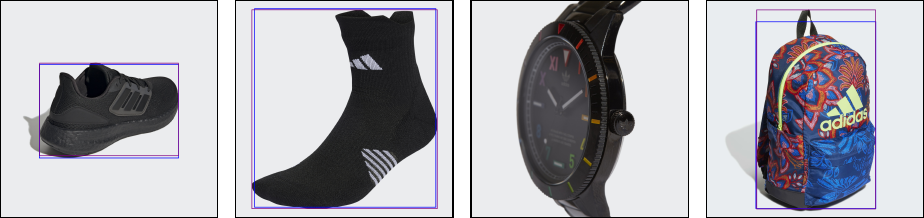}};

        \node[align=left, anchor=north west, inner sep=2pt, fill=black!30!white, text opacity=1, fill opacity=\opacity, text=blue, font=\fontsize{5}{5}\selectfont] at (-4.38,1.03) {hat: 0.91 \\ {\color{violet}top, t-shirt: 0.42}};

        \node[align=left, anchor=north west, inner sep=2pt, fill=black!30!white, text opacity=1, fill opacity=\opacity, text=blue, font=\fontsize{5}{5}\selectfont] at (-2.15,1.03) {dress: 0.99 \\ {\color{violet}dress: 0.86}};

        \node[align=left, anchor=north west, inner sep=2pt, fill=black!30!white, text opacity=1, fill opacity=\opacity, text=blue, font=\fontsize{5}{5}\selectfont] at (2.32,1.03) {jacket: 0.96 \\ {\color{violet}jacket: 0.79}};

        \node[align=left, anchor=east, inner sep=2pt, text=gray, font=\fontsize{5}{5}\selectfont] at (-2.27,-0.93) {2400$\times$2400};
        \node[align=left, anchor=east, inner sep=2pt, text=gray, font=\fontsize{5}{5}\selectfont] at (-.03,-0.93) {2400$\times$2400};
        \node[align=left, anchor=east, inner sep=2pt, text=gray, font=\fontsize{5}{5}\selectfont] at (2.2,-0.93) {2400$\times$2400};
        \node[align=left, anchor=east, inner sep=2pt, text=gray, font=\fontsize{5}{5}\selectfont] at (4.43,-0.93) {2400$\times$2400};
    \end{tikzpicture}
    \vspace*{-7mm}
    \caption{Image examples from \textit{FashionFail-test} and prediction failure cases of two state-of-the-art models on fashion detection, that are \textcolor{blue}{Attribute-Mask R-CNN\textsubscript{SpineNet-143}}\cite{jia2020fashionpedia} and \textcolor{violet}{Fashionformer\textsubscript{Swin-base}}\cite{xu2022fashionformer}.
    }
    \label{fig:ff_samples}
    \vspace{-3mm}
\end{figure}

\begin{figure}
    \centering
    \subfloat[Model predictions at multiple scales. The example image~\cite{zara_fashionpedia} is the same as in \cite{jia2020fashionpedia} and evaluated with the same model.]{
        \begin{tikzpicture}
            \node[inner sep=0pt] at (0.8,0) {\includegraphics[width=0.925\linewidth]{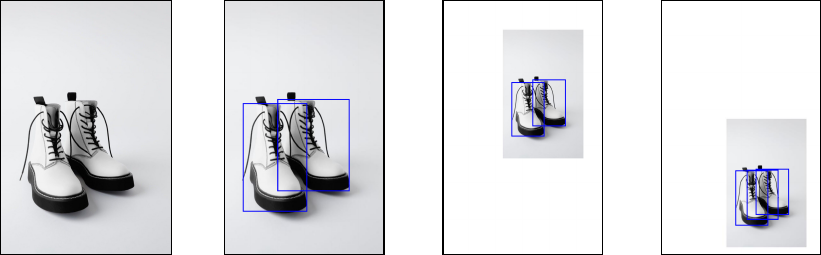}};
            \node[align=left, anchor=west, inner sep=2pt, text=gray, font=\fontsize{5}{5}\selectfont] at (-2.6,-1.15) {1600$\times$1000};
            \node[align=left, anchor=west, inner sep=2pt, text=gray, font=\fontsize{5}{5}\selectfont] at (-0.3,-1.15) {800$\times$500};
            \node[align=left, anchor=west, inner sep=2pt, fill=black!30!white, text=blue, text opacity=1, fill opacity=\opacity, font=\fontsize{5}{5}\selectfont] at (-1.055,1.05) {shoe: 0.52 \\ shoe: 0.28};
            \node[align=left, anchor=west, inner sep=2pt, text=gray, font=\fontsize{5}{5}\selectfont] at (1.7,-1.15) {1600$\times$1000};
            \node[align=left, anchor=west, inner sep=2pt, fill=black!30!white, text=blue, text opacity=1, fill opacity=\opacity, font=\fontsize{5}{5}\selectfont] at (1.125,1.05) { shoe: 0.74 \\ shoe: 0.60};
            \node[align=left, anchor=west, inner sep=2pt, text=gray, font=\fontsize{5}{5}\selectfont] at (4.05,-1.15) {800$\times$500};
            \node[align=left, anchor=west, inner sep=2pt, fill=black!30!white, text=blue, text opacity=1, fill opacity=\opacity, font=\fontsize{5}{5}\selectfont] at (3.31,0.96) { shoe: 0.97 \\ shoe: 0.90 \\ shoe: 0.56};
        \end{tikzpicture}
        \label{fig:amrcnn_zara}
    }
    \hfill
    \subfloat[Model predictions at multiple scales and with different contexts on a sample image~\cite{adidas_watch_context}.]{%
        \begin{tikzpicture}
            \node[inner sep=0pt] at (0,0) {\includegraphics[width=.99\linewidth]{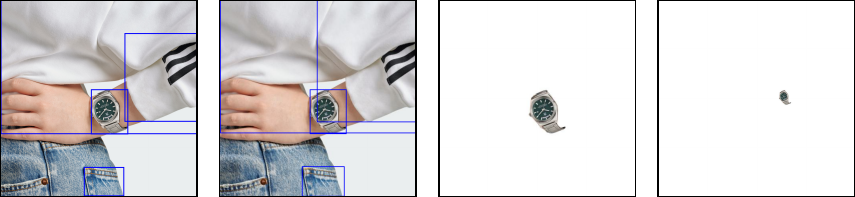}};
            \node[align=left, anchor=west, inner sep=2pt, text=gray, text opacity=1, fill opacity=0.7, fill=black!10!white, font=\fontsize{5}{5}\selectfont] at (-3.425,-0.85) {2400$\times$2400};
            \node[align=left, anchor=west, inner sep=2pt, fill=black!30!white, text=blue, text opacity=1, fill opacity=\opacity, font=\fontsize{5}{5}\selectfont] at (-4.38,0.583) {shoe: 0.94 \\ pocket: 0.84 \\ sleeve: 0.79 \\ top,t-shirt: 0.52};
            \node[align=left, anchor=west, inner sep=2pt, text=gray, text opacity=1, fill opacity=0.7, fill=black!10!white, font=\fontsize{5}{5}\selectfont] at (-0.9835,-0.85) {500$\times$500};
            \node[align=left, anchor=west, inner sep=2pt, fill=black!30!white, text=blue, text opacity=1, fill opacity=\opacity, font=\fontsize{5}{5}\selectfont] at (-2.12,0.587) {watch: 1.00 \\ sleeve: 1.00 \\ top,t-shirt: 0.93 \\ pocket: 0.84};
            \node[align=left, anchor=west, inner sep=2pt, text=gray, font=\fontsize{5}{5}\selectfont] at (1.27,-0.9) {500$\times$500};
            \node[align=left, anchor=west, inner sep=2pt, text=gray, font=\fontsize{5}{5}\selectfont] at (3.33,-0.9) {1500$\times$1500};
        \end{tikzpicture}%
        \label{fig:amrcnn_watch}%
    }

    \caption{The effects of scale and context on bounding box predictions of \textcolor{blue}{Attribute-Mask R-CNN\textsubscript{SpineNet-143}}. The original input images are the first from left. Oversized items and missing context lead to incorrect or non-detections.}
    \label{fig:scale_context_images}
    \vspace{-7mm}
\end{figure}
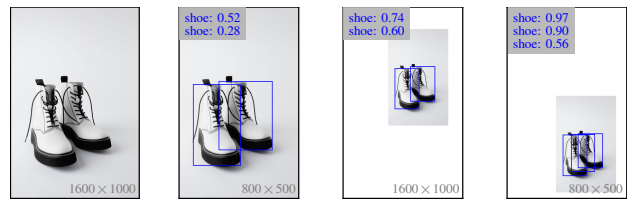
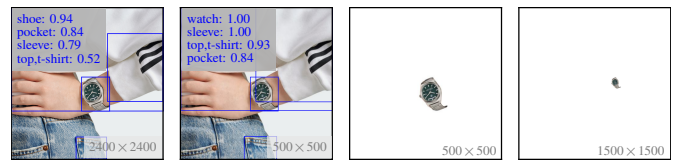



On the contrary, e-commerce fashion images typically display one fashion item on a solid background. As the images of the items are clean, centered, and of high quality, one would expect that the same state-of-the-art models trained on Fashionpedia would perform similarly well on these supposedly simpler e-commerce images. However, in this work we demonstrate that state-of-the-art models face considerable difficulties in processing these clear images, leading to consistent failures. Particularly undesirable among these failure cases are false predictions with overly high confidence or no predictions at all, as depicted in \Cref{fig:ff_samples}.
While it is assumed that models perform ``reasonably well if the apparel item is worn by a model" \cite{jia2020fashionpedia}, pointing out the lack of context as the sole reason for these failures, our findings indicate that the issue is not solely due to the absence of context but also the scale of the apparel items, \cf \Cref{fig:scale_context_images}.
Moreover, \Cref{fig:amrcnn_watch} demonstrates that scale alone may not be sufficient for detection in all cases and that both context and scale are required in some instances. This observation challenges the assumption that such models are well-suited for e-commerce applications and highlights the need for specialized models tailored to unique characteristics of e-commerce fashion images.

Since a robust detection of fashion items is the basis for various fashion applications, we argue that overcoming the mentioned shortcomings would unlock new possibilities for the development of applications within the fashion domain. In particular, addressing these detection failures is essential to advance the development of downstream fashion applications using computer vision.
In addition, given the substantial availability of fashion data, efficient labeling methods within this domain would make an ideal test environment for the development of methods in robust object detection and segmentation that generalize across other domains.

In this work, we introduce the new dataset \emph{FashionFail} along with an efficient data annotation pipeline. The dataset is designed to address the limitations of existing state-of-the-art fashion parsing models. FashionFail consists of a diverse set of 
online shopping images with 
categories that are compatible with the established Fashionpedia \cite{jia2020fashionpedia} ontology. We conduct a thorough evaluation of the detection task on both Fashionpedia and FashionFail with multiple state-of-the-art models.
By providing insights into specific failure modes, our goal is to transparently showcase the limitations of models and emphasize the necessity for further advancements in this research direction.

The paper is structured as follows: in \Cref{sec:related_work}, we provide an overview of the fashion research landscape, outlining key areas, real-world applications, notable datasets, and models. In \Cref{sec:data}, we detail the dataset creation, which covers data scraping, pre-processing, automated annotation and filtering, and quality checks. \Cref{sec:dataset} provides a concise dataset analysis and a comparison with an established existing dataset. Lastly, \Cref{sec:eval} delves into our methodology, evaluation protocol, and an in-depth discussion of the results.

\section{Related Work}\label{sec:related_work}
The existing body of research in the fashion domain 
encompasses synthesis, recommendation, analysis, and detection, 
\cf \cite{cheng2021fashion}.
In the synthesis domain, research has explored topics such as style transfer and pose simulation to create novel fashion compositions~\cite{zhu2023tryondiffusion,kim2023stableviton,he2022style,zhang2023diffcloth}. Fashion recommendation systems~\cite{ding2023computational} have explored outfit compatibility and matching algorithms~\cite{sarkar2023outfittransformer,ye2023show,kang2019complete,zou2022good}. The analysis domain focused on attribute recognition and style learning~\cite{jiao2023learning,han2023fashionsap}. Moreover, detection-based research includes landmark detection, item retrieval, and the task of fashion parsing (combining object detection and segmentation in fashion)~\cite{han2023fame,han2022fashionvil,zhuge2021kaleido}.

For all the mentioned tasks the detection mechanisms form the basis 
enabling the downstream fashion applications~\cite{gu2020fashion}.
In our work, we aim to enhance the effectiveness and reliability of fashion-related tasks by explicitly focusing on fashion parsing, which is an essential aspect of fashion detection.

\textbf{Applications:}
Numerous companies use fashion detection in various real-world applications. Visual Search is a prominent application, evident in platforms such as Amazon's Shop The Look~\cite{stl2022amazon}, Meta's GrokNet~\cite{bell2020groknet}, 
Microsoft's Bing~\cite{hu2018web}, Alibaba~\cite{zhang2018visual}, Ebay~\cite{yang2017visual}, Zalando~\cite{lasserre2019street2fashion2shop}, Pinterest's Complete The Look~\cite{li2020bootstrapping} and Shop The Look~\cite{shiau2020shop}. 
Fashion item retrieval is facilitated by models in Zalando~\cite{lefakis2018feidegger} and NAVER~\cite{kucer2019detect}.
Outfit recommendation systems are implemented by major players such as Amazon~\cite{sarkar2023outfittransformer} and Alibaba~\cite{chen2019pog}, providing users with personalized and stylish outfit suggestions.
Synthesizing fashion is also an active research field \cite{kim2023stableviton, zhu2023tryondiffusion, he2022style}, enabling users to virtually try on clothes before making a purchase.

\textbf{Methods:}
Various methodologies address complex challenges in fashion parsing. Match R-CNN~\cite{ge2019deepfashion2} extends Mask R-CNN to DeepFashion2 \cite{ge2019deepfashion2}, which includes garment detection, landmark estimation, instance segmentation, and consumer-to-shop retrieval. 
Similarly, Attribute-Mask R-CNN~\cite{jia2020fashionpedia} extends Mask R-CNN by an additional prediction head
and introduces a lightweight architecture for effective fashion object detection and segmentation, particularly in attribute prediction. This design mitigates computational demands, presenting a notable alternative in the landscape of fashion parsing methodologies. 
Recently, Fashionformer~\cite{xu2022fashionformer} has been introduced and adopts a distinctive strategy that leverages a single Transformer-based model \cite{dosovitskiy2020image}.
Employing an encoder-decoder framework similar to DETR \cite{carion2020end}, Fashionformer is jointly trained for instance segmentation and attribute recognition. Departing from multi-headed models, Fashionformer achieves state-of-the-art performance on Fashionpedia \cite{jia2020fashionpedia}, establishing itself as the current state-of-the-art in the field.

\textbf{Datasets:}
ModaNet~\cite{zheng2018modanet}, DeepFashion2~\cite{ge2019deepfashion2}, and Fashionpedia~\cite{jia2020fashionpedia} stand out as widely recognized datasets in fashion parsing. However, these datasets primarily consist of ``in the wild" images -- street photos of people wearing clothes. While beneficial for various tasks, such datasets may not perfectly align with the requirements of domains such as e-commerce, which often demand images with clean backgrounds and close-up shots. 
Polyvore Outfits~\cite{vasileva2018learning} or Pinterest CTL~\cite{li2020bootstrapping} 
offer extensive datasets of collage-like images 
presenting outfits that encompass multiple objects. However, this unique format complicates the
analysis of single objects due to the low pixel resolution per object. Additionally, these datasets lack pixel-level object masks, a crucial component for certain applications that require detailed segmentation information.

While existing datasets have been instrumental in advancing fashion research, their limitations are evident.
A significant gap persists, with no datasets specifically tailored for the e-commerce setting, which is characterized by larger, single objects with clean backgrounds and no contextual elements.
In response to these challenges, we present FashionFail, a precisely curated dataset designed to complement the existing dataset landscape. 

\section{Dataset Specification and Collection}\label{sec:data}
Fashion datasets often use images from free-license photo websites, which typically exhibit low resolution. For instance, Fashionpedia \cite{jia2020fashionpedia} training images have an average resolution of $755$ (width) × $986$ (height) pixels.
Consequently, we opted to use images from adidas AG due to their superior image quality and alignment with Fashionpedia categories. 
In the following, we detail the process of data collection and annotation, alongside basic information about the resulting dataset. 

\subsection{Scraping and Filtering}\label{sec:scraping}
\begin{figure}
    \centering
    \captionsetup[subfloat]{labelfont=scriptsize,textfont=scriptsize, farskip=1pt}
        \subfloat{\includegraphics[width=0.32\linewidth]{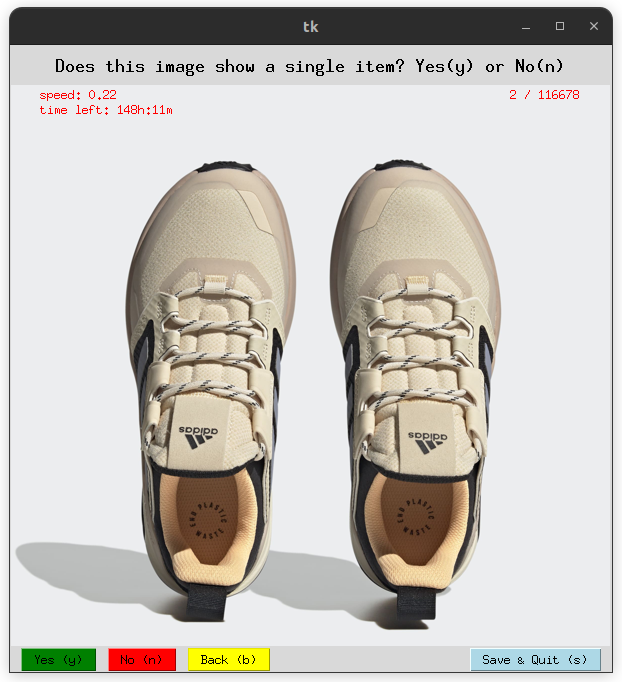}}\hfill
        \subfloat{\includegraphics[width=0.32\linewidth]{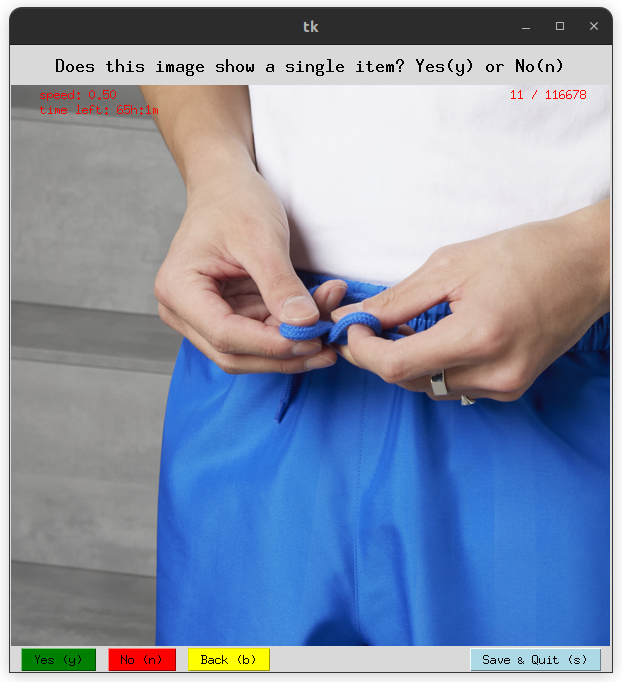}}\hfill
        \subfloat{\includegraphics[width=0.32\linewidth]{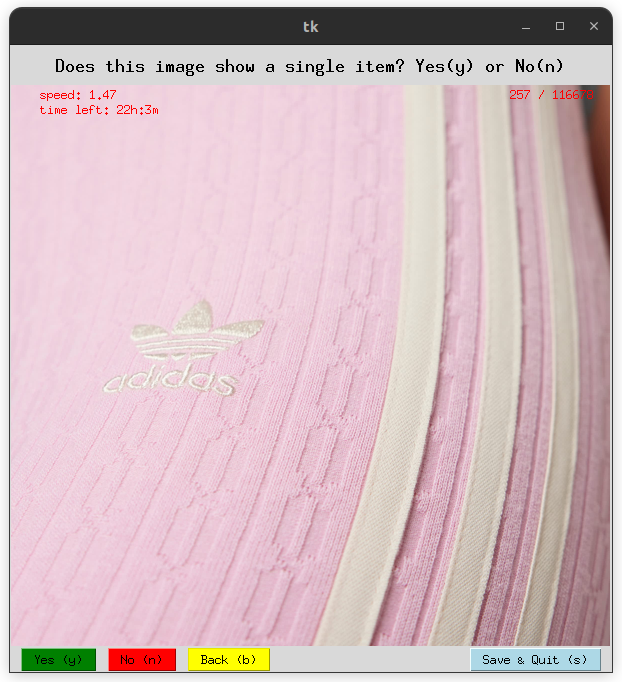}}
    \caption{Screenshots of our filtering tool for three images. Annotators can label images with the click of a button~(bottom left) or use a keyboard shortcut for faster labeling. The top-left displays statistics such as speed~(images per second) and expected time left, while the top-right shows the total number of images to label.}
    \label{appx:fig:labeling_single_crit}
\end{figure}

A custom web crawler built with Scrapy~(\url{https://scrapy.org})
automatically collected over $10,000$ product entries, including images and descriptions, from Adidas website~(\url{https://adidas.de})
. To accelerate the results, we randomly sampled $2,500$ products.
The human annotator then manually filtered images based on three strict criteria: (1) presence of multiple objects or instances of the same object, (2) visibility of any part of the human body, and (3) extreme close-ups hindering category determination from the image, as shown in~\Cref{appx:fig:labeling_single_crit}. 
Images meeting any criterion were excluded to ensure pure, clean, and informative e-commerce product images devoid of contextual information, in contrast to Fashionpedia. The filtering process was efficient, taking less than a second per image using our simple tool, resulting in a total of $5,795$ images.

\begin{figure}
  \centering
   \includegraphics[width=0.99\linewidth, clip, trim=0 1.1cm 0 0]{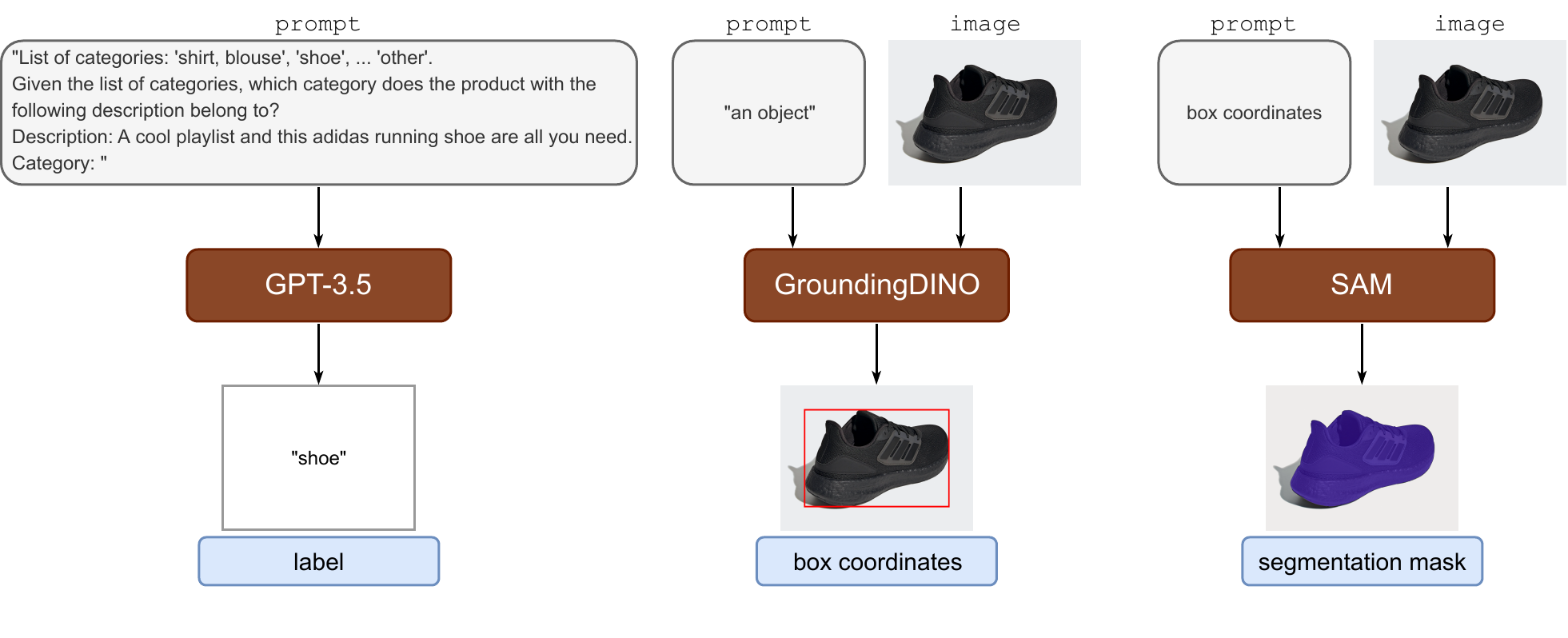}
   \caption{The annotation pipeline employed in curating FashionFail. \textbf{Left:} GPT-3.5 \cite{brown2020language} is prompted with the product description to predict an apparel label. \textbf{Middle:} Grounding DINO \cite{liu2023grounding}, when provided with the product image and a generic text prompt like ``an object", accurately derives bounding box coordinates for all categories. \textbf{Right:} SAM \cite{kirillov2023segment}, in conjunction with the box coordinates and the product image, produces precise segmentation masks.}
   \label{fig:ann_pipeline}
   \vspace{-5mm}
\end{figure}

\subsection{Annotation Pipeline}
Expert annotation is often a laborious and time-consuming task, particularly for pixel-wise label annotations, which are fine-grained and expensive~\cite{ji2018semantic}. Hence, we developed a novel method for automatically annotating our collected data. ~\Cref{fig:ann_pipeline} illustrates the complete pipeline.
We employed GPT-3.5~\cite{brown2020language} (text-davinci-003) to annotate the Fashionpedia category by prompting it with the product description obtained by scraping, as described in \Cref{sec:scraping}.
To annotate the coordinates of the bounding box, we utilized Grounding DINO~\cite{liu2023grounding}, prompting it with the text ``an object'' alongside the product image. Note that we chose a generic text prompt instead of using the category derived in the preceding step as otherwise Grounding DINO would encounter the same difficulties as state-of-the-art fashion parsing models in detecting single fashion items within our collected e-commerce data.
Finally, we employed Segment Anything~(SAM)~\cite{kirillov2023segment} to generate segmentation masks, prompting it with the previously collected bounding box coordinates and the product image. 
Subsequently, any encountered anomaly, such as a sample without a label annotation or multiple box annotations, was automatically filtered, which resulted in a total of $2,682$ images.
This comprehensive approach not only automates the annotation process but also addresses specific challenges associated with different annotation types, resulting in a highly efficient and accurate annotation pipeline.

\begin{figure}
    \centering
    \includegraphics[width=0.99\linewidth]{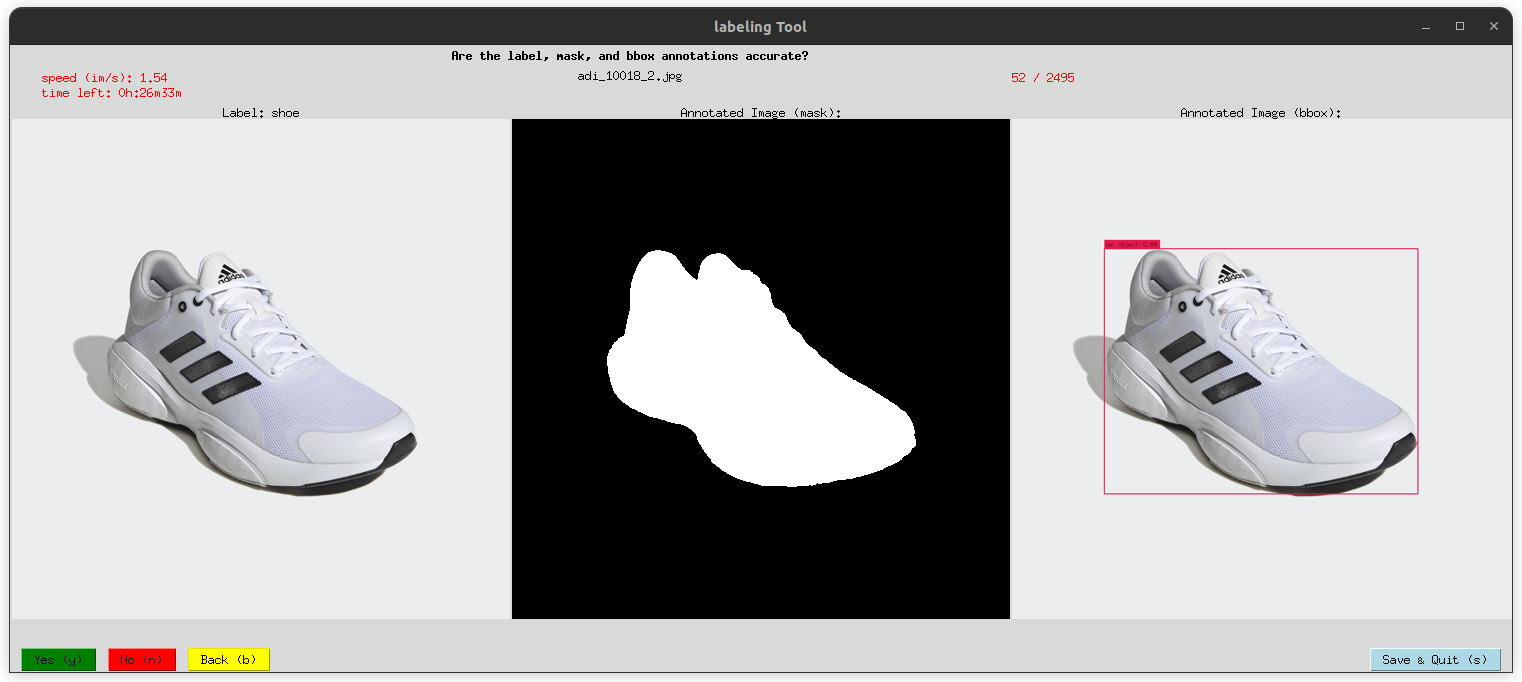}
    \caption{A screenshot of our simple tool used during the data quality review. \textbf{Left:} original image and the label generated by the LLM. \textbf{Middle:} the binary segmentation mask~(white pixels) obtained from SAM. \textbf{Right:} visualization of the bounding box generated by Grounding DINO.}
    \label{appx:fig:quality_review}
    \vspace{-5mm}
\end{figure}

\subsection{Quality Review}
Following the automated stages, a manual quality review is conducted where human annotators use our straightforward interface to flag any invalid samples, as illustrated in~\Cref{appx:fig:quality_review}. These tagged samples are excluded in case of incorrect class labels, or inaccurate box or mask annotations. Human annotators only need to review the automatically annotated samples. This combination of automation and human oversight optimizes the efficiency and accuracy of the annotation process.

  

\subsection{Category Selection}
\begin{figure*}[htbp]
    \centering
    \captionsetup[subfloat]{labelfont=scriptsize,textfont=scriptsize}
    \subfloat[Distribution of apparel categories within each split in both Fashionpedia and FashionFail. (\emph{FashionFail-val} lacks samples for certain categories).]{\includegraphics[width=0.465\textwidth]{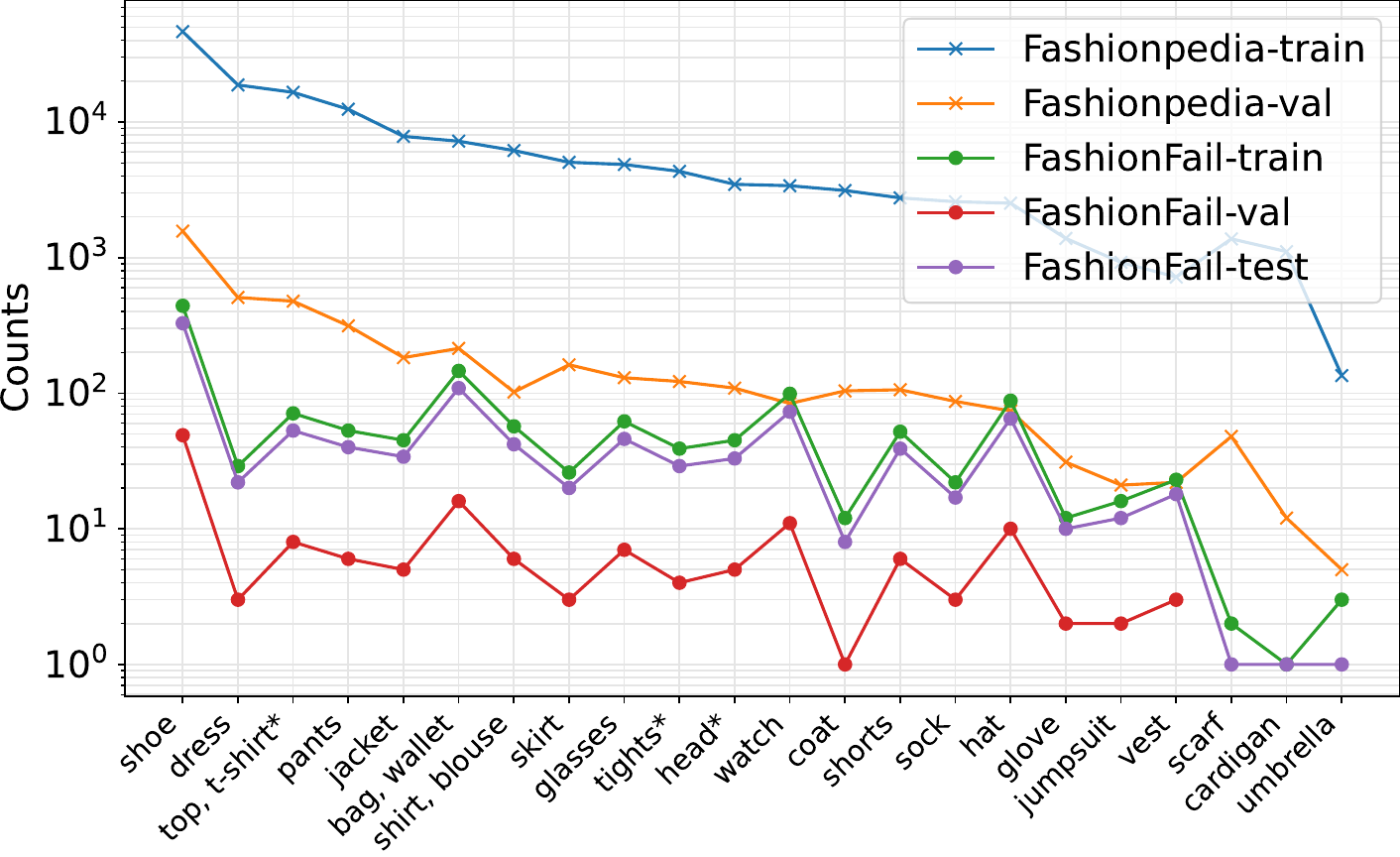}\label{fig:subfigures_dist}}
    \hfill 
    \subfloat[Relative segmentation mask size (square root of mask-area-divided-by-image-area~\cite{gupta2019lvis}) compared between Fashionpedia and FashionFail.]{\includegraphics[width=0.48\textwidth]{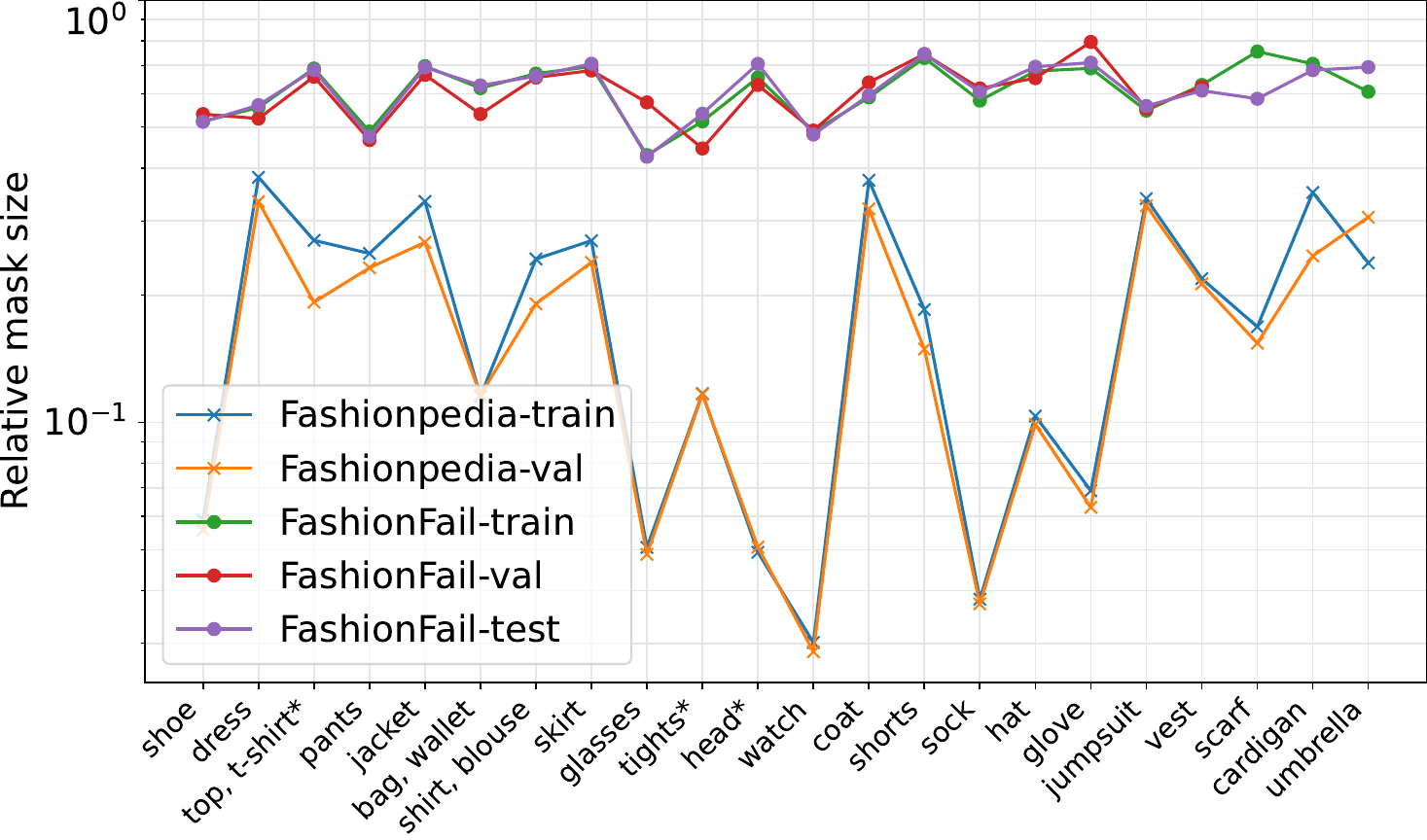}\label{fig:subfigures_mask}}
\caption{Comparison of dataset statistics of Fashionpedia and FashionFail. Best viewed digitally.}
\label{fig:subfigures}
\end{figure*}

FashionFail is aligned with practical applications in the fashion domain, particularly in clothing recommendation systems and e-commerce, where primary items play a central role. Therefore, the dataset focuses on primary clothing, such as jackets, pants, and shoes. In alignment with the Fashionpedia ontology, categories falling under `garment parts', `closures', and `decorations' super-categories are excluded. This eliminates categories such as \textit{sleeve, pocket, applique}, and others. Additionally, due to an insufficient number of images, categories such as \textit{sweater, cape, tie, belt}, and \textit{leg warmer} were removed. The refined dataset comprises 22 categories, outlined in~\Cref{fig:subfigures}.

\section{Dataset Statistics}\label{sec:dataset}
Through our automated annotation pipeline and careful review process, each image within FashionFail is precisely annotated with a single label, bounding box, and segmentation mask. This process has resulted in a dataset comprising $2,495$ pairs of images and annotations.

The dataset is divided into training, validation, and test sets, consisting of $1,344$, $150$, and $1,001$ images, respectively. The split preserves class frequencies in each subset, and maintains a similar distribution to Fashionpedia. \Cref{fig:subfigures_dist} details the class distribution in different splits. For example, the most common class in both datasets is `shoe', constituting $30\%$ in Fashionpedia and $33\%$ in FashionFail, respectively.

All FashionFail images maintain a consistent resolution of $2400$ × $2400$ pixels, representing a considerable improvement compared to Fashionpedia's resolution of $755$ × $986$. 
Each FashionFail image exclusively features a singular item, centered within the frame. Despite its likewise object-centered distribution, Fashionpedia consists of smaller objects in terms of relative size, as depicted in~\Cref{fig:subfigures_mask}. With the heightened resolution and larger relative object size, FashionFail offers significantly finer pixel-level masks for its objects.

\section{Experiments, Evaluation and Results}\label{sec:eval}

In this section, we outline the experiments conducted, covering model architectures, training schedules, data augmentations, evaluation, and other relevant details, followed by a comprehensive discussion of the results.

\subsection{Experiments}
In our experimentation, we employed two top-performing models on the Fashionpedia dataset, alongside our Mask R-CNN based models. 
To facilitate a fair and unbiased comparison between models, we included variations of the two top-performing models that share the same backbone network as our Mask R-CNN.

\textbf{Attribute-Mask R-CNN~(A-MRCNN)}~\cite{jia2020fashionpedia}
is trained for instance segmentation with attribute localization on Fashionpedia.
Their best-performing model
incorporates SpineNet-143~\cite{du2020spinenet} as backbone and adheres to a training schedule of $6\times$ (189 epochs of training) with a linear scaling of the learning rate and a batch size of 256. Notably, during training, the model benefits from large-scale jittering in addition to standard random horizontal flipping and cropping. For a comprehensive comparison, we also present results for A-MRCNN using the ResNet-50~\cite{he2016deep} backbone with a feature pyramid network~(R50-FPN)~\cite{lin2017feature}.
We use the publicly available official implementation\footnote{\url{https://bit.ly/GitHub-Fashionpedia}}.

\textbf{Fashionformer}~\cite{xu2022fashionformer}
is the current state-of-the-art model on the Fashionpedia dataset. Unlike previous work that treats each task separately as a multi-head prediction problem, Fashionformer uses a single unified 
vision transformer \cite{dosovitskiy2020image} for multiple tasks. Their best-performing model uses Swin-Transformer~\cite{liu2021swin} as its backbone. They adopted $3\times$ schedule (95 epochs of training) with a batch size of 16 while using large-scale jittering. For a comprehensive comparison, we also present results for Fashionformer using R50-FPN as backbone.
We use the publicly available official implementation\footnote{\url{https://github.com/xushilin1/FashionFormer}}.

\begin{figure}
    \centering
    \captionsetup[subfloat]{farskip=3pt,captionskip=1pt}
    \subfloat{\includegraphics[width=0.48\textwidth]{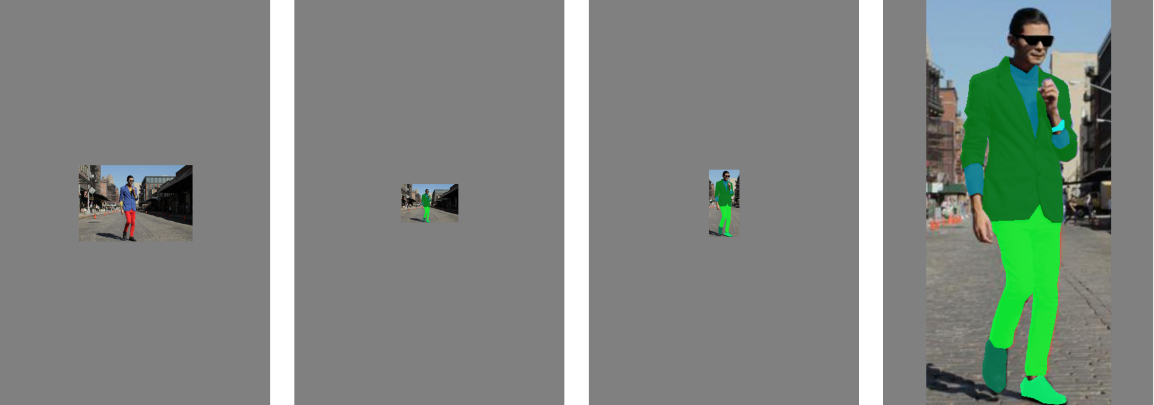}\label{fig:aug_2}}
    \hfill
    \subfloat{\includegraphics[width=0.48\textwidth]{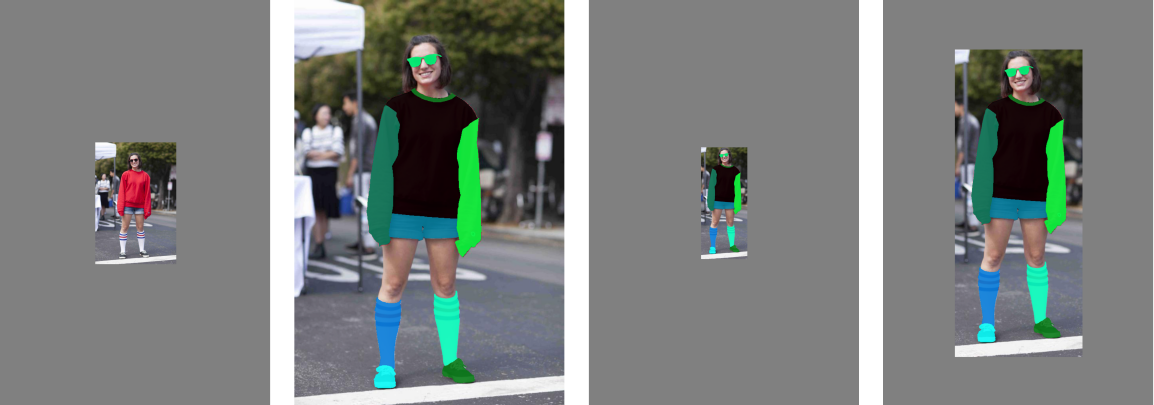}\label{fig:aug_1}}
    \hfill
    \subfloat{\includegraphics[width=0.48\textwidth]{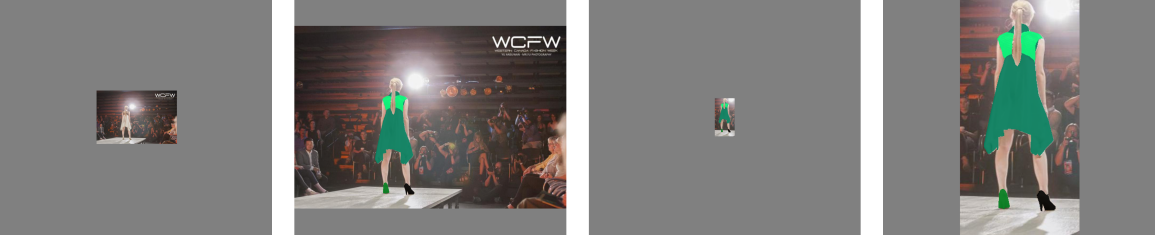}\label{fig:aug_3}}
\caption{Illustration of data augmentations of Fashionpedia training samples employed during training of our Facere model. From left to right: (1) original images, (2) large-scale jittering (resize range $\in [0.1, 2]$), (3) custom box crop, and (4) jittering and cropping 
combined. 
Note that in this illustration the displayed examples are padded (gray area) just to visualize scale variations.}
\label{fig:augmentation_examples}
\end{figure}

\textbf{Facere\footnote{Facere 
is Latin for `to make', from which the word fashion is derived.}}
is based on Mask R-CNN~\cite{li2021benchmarking} with a R50-FPN backbone, which has been initially pretrained on the COCO dataset~\cite{lin2014microsoft}. We finetuned Facere using a modified 75-25 split of \textit{Fashionpedia-train}, where 75\% used for training and 25\% for validation, for 125 epochs~($\approx3\times$ schedule) with a batch size of 8 and Adam~\cite{kingma2014adam} as optimizer with default parameters from PyTorch. 
For faster training and reduced memory usage, we used half-precision (16-bit) training provided by PyTorch Lightning\footnote{\url{https://lightning.ai/docs/pytorch/}}, converting it back to full precision (32-bit) for enhanced performance.
Our training incorporated standard techniques such as horizontal flipping, photometric distortion~\cite{liu2016ssd}, and large-scale jittering~\cite{ghiasi2021simple}. To deal with the varying scale of items while minimizing context information from the background, we additionally introduce a simple yet effective data augmentation approach: 
we crop the image based on the smallest box containing all bounding boxes (inspired by 
\cite{taesiri2023zoom}) and apply large-scale jittering to these custom image crops. All data augmentation techniques are applied with a probability of 50\%. Examples of the augmentations are illustrated in \Cref{fig:augmentation_examples}.

\textbf{Facere+} follows the same setup as Facere 
but finetuned on \textit{FashionFail-train} for 150 epochs.
The motivation behind Facere+ is to showcase the learnability of FashionFail-- a form of data quality check 
for the usability and cleanliness of FashionFail.

\begin{table*}[htbp]
\setlength{\tabcolsep}{5pt}
\caption{Results on the \textit{FashionFail-test} dataset assessed through a range of metrics from \Cref{sec:eval_metrics}.}
\label{tab:eval_metrics_ff} \centering
    \begin{tabular}{l||c|c|c||c|c|c|c|c|c|c|c|c|c}
    \toprule
    & & & \textbf{FashionFail} & \multicolumn{2}{c|}{$\mathrm{mAP}_w$}  & \multicolumn{2}{c|}{$\mathrm{mAP}_w^{@\mathrm{IoU}=.50}$} & \multicolumn{2}{c|}{$\mathrm{mAP}_w^{@\mathrm{IoU}=.75}$} & \multicolumn{2}{c|}{$\mathrm{mAR}_w^{\mathrm{top}1}$} & \multicolumn{2}{c}{$\mathrm{mAR}_w^{\mathrm{top}100}$} \\ 
    \textbf{Method} & \textbf{Backbone} & \textbf{Params} & \textbf{train data} & \multicolumn{1}{c}{box} & \multicolumn{1}{c|}{mask} & \multicolumn{1}{c}{box} & \multicolumn{1}{c|}{mask} & \multicolumn{1}{c}{box} & \multicolumn{1}{c|}{mask} & \multicolumn{1}{c}{box} & \multicolumn{1}{c|}{mask} & \multicolumn{1}{c}{box} & \multicolumn{1}{c}{mask} \\
    \toprule
    A-MRCNN ~\cite{jia2020fashionpedia} & \text{SpineNet-143} & 79.2M & \xmark
& 21.0 & \underline{21.2}  & 22.1 & 22.0  & \underline{21.5} & \underline{21.5}   & 32.1 & \underline{32.1}  & 32.7 & 32.7 \\
    Fashionformer~\cite{xu2022fashionformer} & Swin-base &  106.8M & \xmark
& \underline{22.2} & 20.9  & \underline{25.1} & \underline{23.2}   & \underline{21.5} & 20.2   & \underline{32.9} & 30.2  & \underline{44.6} & \underline{40.0} \\ \midrule
    A-MRCNN~\cite{jia2020fashionpedia} & ResNet50-FPN & 46.4M & \xmark
& 18.3 & 18.6 & 19.7 & 19.2 & 19.3 & 19.0 & 25.7 & 25.6 & 25.8 & 25.6 \\
    Fashionformer~\cite{xu2022fashionformer} & ResNet50-FPN &  43.8M & \xmark
& 14.4 & 14.5 & 14.6 & 14.6 & 14.6 & 14.6 & 25.9 & 25.9 & 26.9 & 26.3 \\
    Ours: Facere  & ResNet50-FPN & 45.9M & \xmark
& \textbf{24.0} & \textbf{24.3}  & \textbf{27.5} & \textbf{26.8}   & \textbf{25.1} & \textbf{24.9}   & \textbf{44.6} & \textbf{45.5}  & \textbf{47.5} & \textbf{47.9} \\ \midrule\midrule
Facere+  & ResNet50-FPN & 45.9M & \cmark
& 93.4 & 94.1  & 95.7 & 95.7   & 95.3 & 95.6   & 96.6 & 97.3  & 96.6 & 97.3 \\
    \bottomrule
    \end{tabular}
\end{table*}
\begin{table}[htbp]
\setlength{\tabcolsep}{2.5pt}
\caption{Results for the same evaluation as in \Cref{tab:eval_metrics_ff} but on the \textit{Fashionpedia-val} dataset. Facere maintains similar performance to other ResNet50-FPN based models.}
\label{tab:eval_metrics_fp} \centering
    \begin{tabular}{l||c|c|c|c|c|c|c|c|c|c}
    \toprule
    \textbf{Method} & \multicolumn{2}{c|}{$\mathrm{mAP}_w$}  & \multicolumn{2}{c|}{$\mathrm{mAP}_w^{@.50}$} & \multicolumn{2}{c|}{$\mathrm{mAP}_w^{@.75}$} & \multicolumn{2}{c|}{$\mathrm{mAR}_w^{1}$} & \multicolumn{2}{c}{$\mathrm{mAR}_w^{100}$} \\ \toprule
    A-MRCNN
& \underline{66.6} & 60.1 & \underline{84.3} & 82.0  & \underline{73.7} & 67.1  & 65.2 & 60.9  & 75.5 & 69.0 \\
    Fashionformer
& \textbf{72.4} & \textbf{70.6} & \textbf{87.6} & \textbf{90.0}  & \textbf{77.9} & \textbf{77.3}  & \textbf{71.1} & \textbf{69.1}  & \textbf{81.9} & \textbf{78.4} \\ \midrule
    A-MRCNN
& 64.3 & 58.3 & 83.2 & 80.6  & 72.7 & 64.9  & 63.4 & 60.0  & 73.4 & 67.6 \\
    Fashionformer
& 64.4 & \underline{64.2} & 82.4 & \underline{85.0}  & 69.3 & \underline{69.9}  & \underline{65.8} & \underline{65.0}  & \underline{77.5} & \underline{74.6} \\
    Facere
& 63.9 & 58.0 & 82.5 & 80.3  & 71.6 & 64.3  & 64.5 & 60.8  & 76.0 & 69.4 \\
    \bottomrule
    \end{tabular}
\end{table}
\subsection{Evaluation Metrics}\label{sec:eval_metrics}

A prediction is called a true positive (TP) if it matches the ground truth class and exceeds a specified intersection over union (IoU) \cite{Jaccard1912THEDO} threshold; otherwise, it is called a false positive (FP). Similarly, a ground truth object is identified as a false negative (FN) if there are no predictions for that object that surpass the specified IoU detection threshold.
Furthermore, object detectors assign a (normalized) score to each predicted bounding box~(referred to as confidence score), indicating the likelihood of the box containing an object of a specific class. Typically, boxes with low confidence scores are rejected. In our evaluation, we maintained the default of $0.05$ as the minimum confidence threshold.
The quantities TP, FP, and FN are used to determine measures such as $\mathrm{precision}=\frac{\mathrm{TP}}{\mathrm{TP}+\mathrm{FP}}$ and $\mathrm{recall}=\frac{\mathrm{TP}}{\mathrm{TP}+\mathrm{FN}}$. Note that they also depend on the specified IoU threshold $\phi$ and confidence threshold $\tau$. In the following, we use the notation $\mathrm{precision}_\phi(\tau)$ and $\mathrm{recall}_\phi(\tau)$ to explicitly denote the dependency of the measures on the thresholds.

The interpolated average precision~(AP)~\cite{salton1986} is a commonly used metric to assess object detection performance, which is defined as the mean precision at a set of equally spaced recall levels, \eg for 101 points $\mathcal{R} = \{ 0.00, 0.01, 0.02, \ldots, 1.00 \}$:
\begin{align}\label{eq:interpol_ap}
\begin{split}
    \mathrm{AP}^{@\mathrm{IoU}=\phi} = \frac{1}{|\mathcal{T}_\phi|} \sum_{\tau\in T_\phi} \max \{ &\text{precision}_\phi(t) : t\in \mathcal{T}_\phi, \\
    & \mathrm{recall}_\phi(t)\geq \mathrm{recall}_\phi(\tau) \}
\end{split}
\end{align}
where $\mathcal{T}_\phi:=\{t \in [0,1]: \mathrm{recall}_\phi(t)\in \mathcal{R}\}$.\footnote{Note that AP metrics, as employed by Pascal VOC~\cite{everingham2010pascal} and MS COCO~\cite{lin2014microsoft}, feature slightly different calculations. 
} 


However, this metric might not be ideal for small datasets, particularly those that are unbalanced, akin to FashionFail.
Consider the ``umbrella" class in our \textit{FashionFail-test}, with only one sample among $1,001$ examples. If a model accurately predicts this single sample, it could significantly inflate the mean of average precision (mAP) across classes. 
Therefore, in our evaluation, we use a weighted version of mAP \cite{zhao2015deep}:
\begin{equation}\label{eq:map_w}
    \mathrm{mAP}^{@\mathrm{IoU}=\phi}_w = \frac{1}{C} \sum_{c=1}^{C}w_c \cdot \mathrm{AP}^{@\mathrm{IoU}=\phi}_c
\end{equation}
where the weight $w_c$ is the relative frequency of class $c \in \{ 1,\ldots,C\}$ in the test set
and $\mathrm{AP}^{@\mathrm{IoU}=\phi}_c$ is the interpolated AP according to (\ref{eq:interpol_ap}) for class $c$.
To have a threshold-free evaluation metric, the weighted mAP from (\ref{eq:map_w}) is additionally averaged over a range of IoU thresholds $\mathcal{I} = \{ 0.50, 0.55, \ldots 0.95\}$:
\begin{equation}\label{eq:map}
    \mathrm{mAP}_w = \frac{1}{|I|} \sum_{\phi \in \mathcal{I}} \mathrm{mAP}_w^{@\mathrm{IoU}=\phi}
\end{equation}
which also serves as our primary evaluation metric.

Analogously, we also report the recall for a fixed number of maximum detections $k \in \mathbb{N}$. Given the set of all confidences $\mathcal{S}_{n}$ of the predictions for test example $n \in \{1,\ldots,N\}$, where $\tau_1, \ldots, \tau_{|\mathcal{S}_{n}|} \in \mathcal{S}_{n}$ are assumed to be in decreasing order, the average recall for $k$ detections is defined as 
\begin{equation}\label{eq:average_recall}
    \mathrm{AR}^{\mathrm{top}\,k} = \frac{1}{N\,|\mathcal{I}|} \sum_{n=1}^N \sum_{\phi \in \mathcal{I}} \mathrm{recall}_\phi(\tau;n)\!\!\restriction_{\tau \in \min\{\mathcal{S}^k _{n} \cup\, \{1.0\} \} } 
\end{equation}
with $\mathcal{S}_{n}^k = \{\tau_j \in \mathcal{S}_{n}: j = 1,\ldots, \min\{k, |\mathcal{S}_{n}|\} \}$ if $|\mathcal{S}_{n}|>0$, else $\mathcal{S}_{n}^k = \mathcal{S}_{n}^k = \emptyset$, the set of the top $k$ confidences in test example $n$ and $\mathrm{recall}_\phi(\boldsymbol{\cdot};n)$ the recall score only for test example $n$. Thus, the weighted mean average recall \cite{hosang2015makes} for the top $k$ detections is computed as
\begin{equation}
    \mathrm{mAR}_w^{\mathrm{top}\,k} = \frac{1}{C} \sum_{c=1}^C w_c  \cdot \mathrm{AR}^{\mathrm{top}\,k}_c
\end{equation}
where the class weights $w_c$ for classes $c \in \{1,\ldots,C\}$ are the same as in \Cref{eq:map_w} and $\mathrm{AR}^{\mathrm{top}\,k}_c$ is the average recall as in \Cref{eq:average_recall} but restricted on class $c$.

\subsection{Results and Analysis}\label{sec:results-analysis}

In~\Cref{tab:eval_metrics_ff}, we present the model performance on the \textit{FashionFail-test} dataset, evaluated using several metrics covering both detection and segmentation. Facere significantly outperforms state-of-the-art models, even those with larger backbones, across all evaluation metrics. It exhibits improvement in box-$\text{mAP}_{\text{w}}$ over Fashionformer by $+1.8$ and mask-$\text{mAP}_{\text{w}}$ over A-MRCNN by $+3.1$. Furthermore, as the box-IoU increases, indicating a stricter model, the performance gap between Facere and the second best-performing model widens from $+2.4$ to $+3.6$. Remarkably, Facere outperforms all models on both box- and mask-$\text{AR}^{\text{top1}}$, surpassing the closest competitor by $+11.7$ and $+13.4$, respectively.
The performance gap in $\text{AR}^{\text{top100}}$ diminishes due to the enhanced performance of Swin-base Fashionformer. This improvement stems from Fashionformer generating 100 predictions for each image without employing filtering techniques such as non-maximum suppression. However, the R50-FPN counterpart of Fashionformer does not experience as much benefit. Nevertheless, Facere maintains its superiority over all models. 
Lastly, Facere+, which used \textit{FashionFail-train} during training, shows exceptional performance across all metrics, indicating that the constructed dataset is adequate. 

In~\Cref{tab:eval_metrics_fp}, we present the model performance on the \textit{Fashionpedia-val} dataset, evaluated using the same metrics. As expected, Swin-base Fashionformer outperforms other models. In addition, Facere achieves comparable results to its R50-FPN counterparts, with insignificant performance differences of $-0.5$ for box-$\text{mAP}_{\text{w}}$ and $-1.3$ for box-$\text{mAR}_{\text{w}}$.
Our objective is not to improve the performance of our models on \textit{Fashionpedia-val} but to improve 
the robustness to other datasets or domains, such as \textit{FashionFail-test}, while maintaining comparable performance to the baselines on \textit{Fashionpedia-val}.
With negligible performance differences between Facere and its R50-FPN counterparts, we observe that Facere effectively aligns with our goals by maintaining on-par performance on its training data while enhancing 
robustness against scale and context.



\begin{figure}
    \centering
    \includegraphics[width=.99\linewidth]{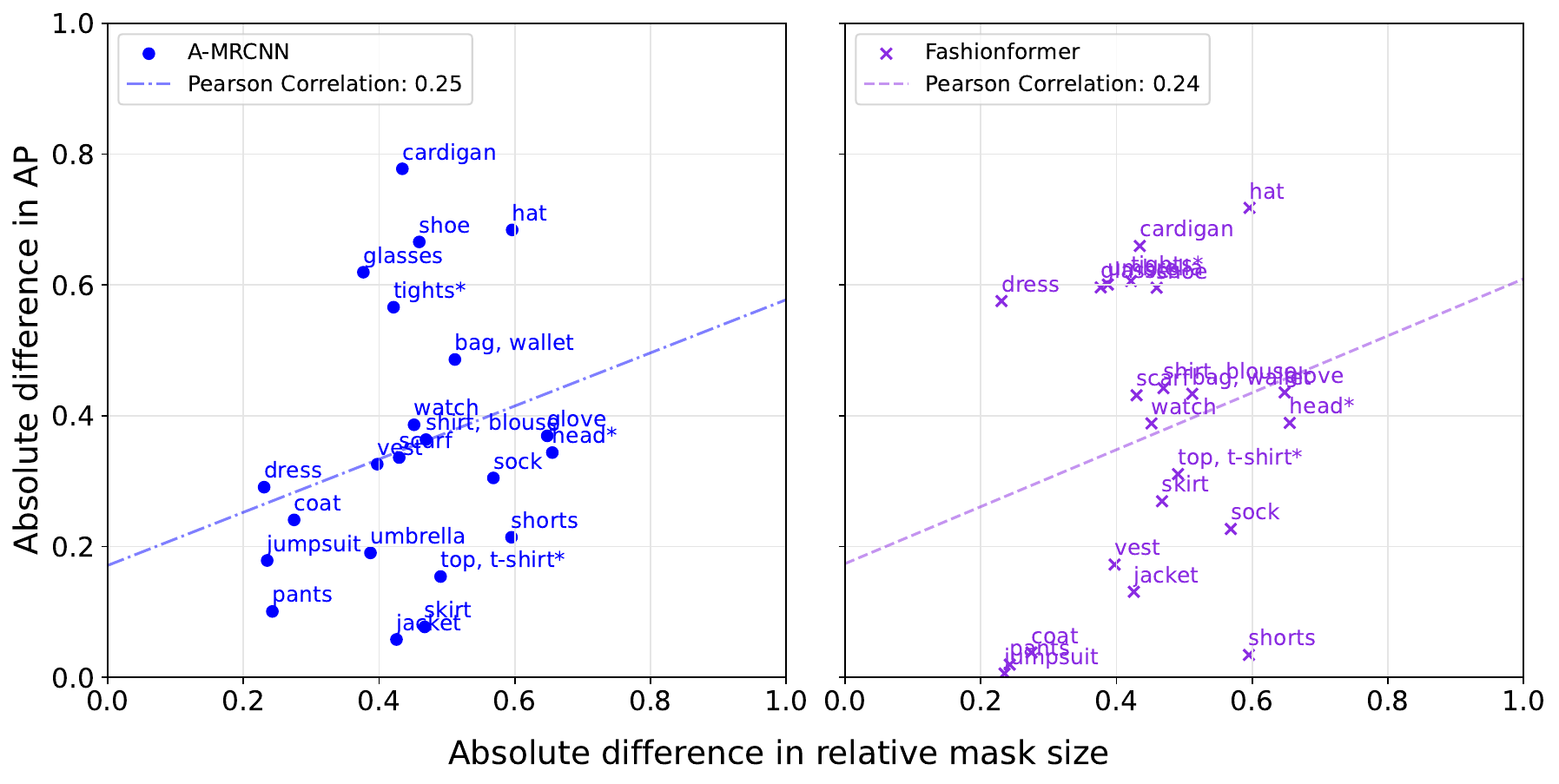}
    \caption{Correlation analysis between the relative sizes of objects in Fashionpedia and FashionFail and the performance of \textcolor{blue}{A-MRCNN\textsubscript{R50-FPN}} and \textcolor{violet}{Fashionformer\textsubscript{R50-FPN}} on both datasets. The observed positive correlation indicates that the greater the difference in relative object size, the greater the difference in model performance between the two datasets.}
    \label{fig:scale_diff_scatter}
\end{figure}


\begin{figure*}[htbp]
    \centering
    \captionsetup[subfloat]{labelfont=scriptsize,textfont=scriptsize}
    \subfloat[Class-wise performance on \textit{Fashionpedia-val}.]{\includegraphics[width=0.34\textwidth]{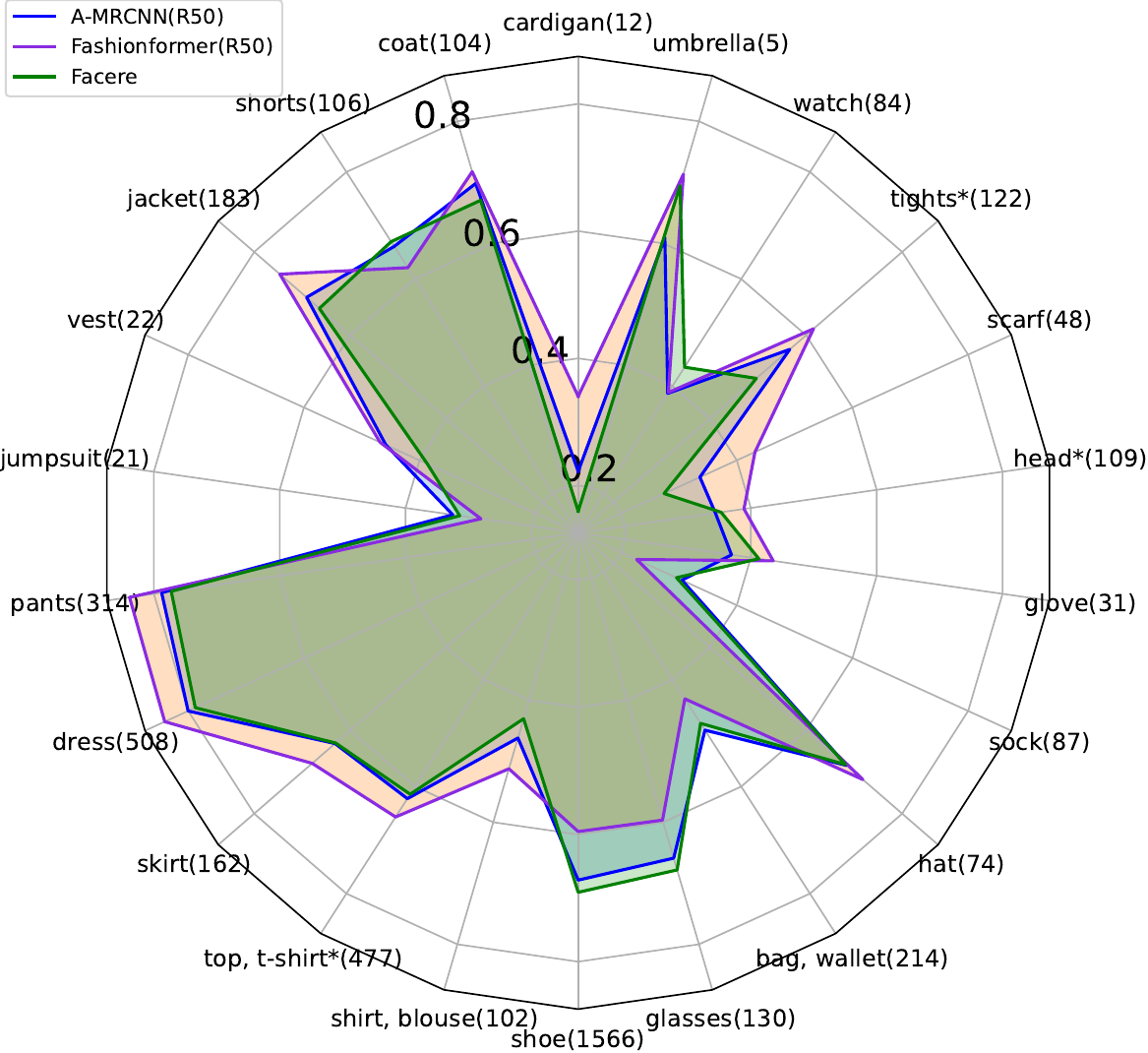}  \label{fig:ap_per_class_fp}}%
    \subfloat[Class-wise performance on \textit{FashionFail-test}.]{\includegraphics[width=0.34\textwidth]{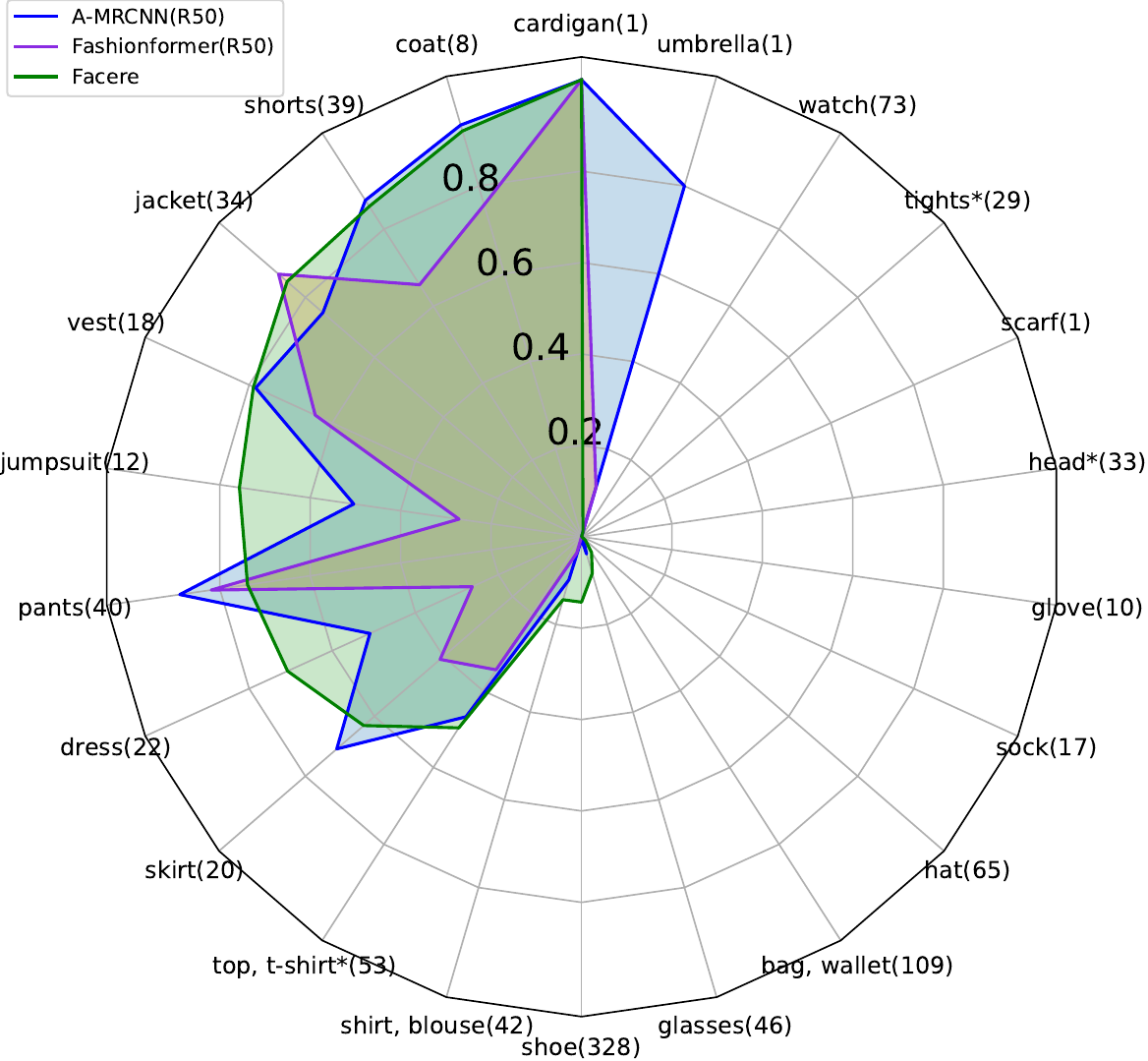}  \label{fig:ap_per_class_ff}}%
    \subfloat[Predictions for \textbf{left}: A-MRCNN\textsubscript{R50-FPN}, \textbf{middle}: Fashionformer\textsubscript{R50-FPN} \textbf{right}: Facere]{\shortstack{
    \includegraphics[width=0.096\textwidth]{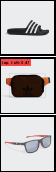} \hfill \includegraphics[width=0.096\textwidth]{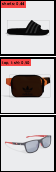} \hfill \includegraphics[width=0.096\textwidth]{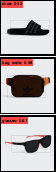} \label{fig:inference_examples}
    }}
\caption{Visualization of box-AP~(averaged over ten IoU thresholds as in (\ref{eq:map})) per class for each model on (a) \textit{Fashionpedia-val} and (b) \textit{FashionFail-test}. Values in brackets represent the number of samples for each category. (c) visualizes the model predictions on \textit{FashionFail-test} with a score threshold of $0.35$. Best viewed digitally.}
\label{fig:ap_per_class}
\end{figure*}

Moreover, we evaluate the model performance at the class level.
To get an intuition about the failure cases, we first analyzed whether there is a correlation between the size of objects in Fashionpedia and the model performance of A-MRCNN and FashionFormer on FashionFail. 
To this end, we calculated the absolute difference in relative mask sizes of objects between \textit{Fashionpedia-val} and \textit{FashionFail-test}, and plotted it against the absolute difference in average precision (AP), see \Cref{fig:scale_diff_scatter}. 
Interestingly, the plots reveal a clear positive correlation between the absolute difference in relative mask size and the absolute difference in AP. 

In \Cref{fig:ap_per_class_fp} and \Cref{fig:ap_per_class_ff} we report the model performances on Fashionpedia and FashionFail per class.
The models exhibit similar performance distribution across classes on \textit{Fashionpedia-val}. 
In particular, Fashionformer performs slightly better on classes such as \textit{dress}, and \textit{skirt}, while Facere shows a slight advantage on the \textit{shoe} class, which is the most dominant class. 
On the other hand, all models perform poorly for most classes on our \textit{FashionFail-test} data. 
However, for classes such as \textit{coat, jacket, pants}, and \textit{dress} the models demonstrate promising results. Notably, Facere performs consistently well in those classes whereas the other models 
generally show inconsistent performances across classes.
Moreover, Facere accurately predicts classes such as \textit{shoe, glasses}, and \textit{bag, wallet}, where the others entirely fail.

These results, combined with the scale analysis from \Cref{fig:scale_diff_scatter}, explain why the models do not suffer from a severe drop in performance on certain classes such as \textit{coat, pants, jackets}, as their mask sizes are similar in both datasets. On the contrary, for classes such as \emph{hat, shoe} the relative object sizes differ significantly between Fashionpedia and FashionFail, leading to a dramatic performance drop on \textit{FashionFail-test}. 

We provide a qualitative comparison of model predictions of A-MRCNN, FashionFormer, and Facere in \Cref{fig:inference_examples}.

\section{Conclusions}\label{sec:conclusion}

In this work, we introduce FashionFail, a dataset for precise fashion object detection and segmentation, serving as a robustness benchmark for fashion parsing models trained on ``in-the-wild'' images such as in Fashionpedia. Our dataset was efficiently created with a novel web crawling and annotation pipeline. Moreover, we address the limitations that state-of-the-art models face in generalizing to the domain of e-commerce images due to scale and context issues. Emphasizing the impact of carefully designed domain-specific data augmentations, we demonstrate their effectiveness in improving model generalizability while preserving original domain performance with our Facere model.
The data augmentations used in Facere provide a simple yet effective approach to improving the robustness of fashion parsing for use in real-world applications. In particular, our results show that even with constraints such as less training data, reduced model complexity, and shorter training times, our approach has the potential to outperform existing and well-established models. 
However, it is important to acknowledge that the presented results still require further refinement.

Moving forward, possible improvements lie in enhancing the efficiency of the LLM within the annotation pipeline, considering the latest state-of-the-art LLMs, and exploring vision-language models. 
In terms of methodology, testing scale-invariant approaches \cite{li2019scale,zhang2022delving} could alleviate the observed scale variation problems.
Furthermore, there is potential to explore the reliability of these models in more detail with further evaluation schemes to test for robust performance.

\bibliographystyle{ieeetr}
\bibliography{bibliography}


\end{document}